\documentclass{article}

\usepackage{ijcai16}

\usepackage{times}
\usepackage{epsfig}
\usepackage{graphicx}
\usepackage{amsmath}
\usepackage{amssymb}
\usepackage{mathsymb}
\usepackage{textcomp}
\usepackage{verbatim}
\usepackage{subfigure}
\usepackage{url}
\usepackage{graphicx}
\usepackage{multirow}
\usepackage[super]{nth}
\usepackage{balance}
\usepackage{colortbl}
\graphicspath{{./}{./Fig/}{./Figs/}{./Fig/eps/}{./Fig/pdf/}}

\usepackage{algorithm2e}

\let\savedalgorithm\algorithm
\let\savedendalgorithm\endalgorithm

\usepackage{algorithm}

\newcommand{\etal}{\textit{et al.\ }}
\newcommand{\eg}{\emph{e.g.,\ }}
\newcommand{\ie}{\emph{i.e.,\ }}

\title{What-and-Where to Match: Deep Spatially Multiplicative Integration Networks for Person Re-identification}
\author{Lin Wu$^{\dag}$$^{\sharp}$, Yang Wang$^{\sharp}$, Xue Li$^{\dag}$, Junbin Gao$^{\S}$\\
 $^{\dag}$ The University of Queensland, Australia\\
 $^{\sharp}$The University of New South Wales, Kensington, Sydney, Australia\\
 $^{\S}$The University of Sydney, Australia
}

\begin{document}

\maketitle

\begin{abstract}
Matching pedestrians across disjoint camera views, known as person re-identification (re-id), is a challenging problem that is of importance to visual recognition and surveillance. Most existing methods exploit local regions within spatial manipulation to perform matching in local correspondence. However, they essentially extract \emph{fixed} representations from pre-divided regions for each image and perform matching based on the extracted representation subsequently. For models in this pipeline, local finer patterns that are crucial to distinguish positive pairs from negative ones cannot be captured, and thus making them underperformed. In this paper, we propose a novel deep multiplicative integration gating function, which answers the question of \emph{what-and-where to match} for effective person re-id. To address \emph{what} to match, our deep network emphasizes common local patterns by learning joint representations in a multiplicative way. The network comprises two Convolutional Neural Networks (CNNs) to extract convolutional activations, and generates relevant descriptors for pedestrian matching. This thus, leads to flexible representations for pair-wise images. To address \emph{where} to match, we combat the spatial misalignment by performing spatially recurrent pooling via a four-directional recurrent neural network to impose spatial dependency over all positions with respect to the entire image. The proposed network is designed to be end-to-end trainable to characterize local pairwise feature interactions in a spatially aligned manner. To demonstrate the superiority of our method, extensive experiments are conducted over three benchmark data sets: VIPeR, CUHK03 and Market-1501.
\end{abstract}


\section{Introduction}\label{sec:intro}
\begin{figure}[t]
\centering
\includegraphics[height=4cm]{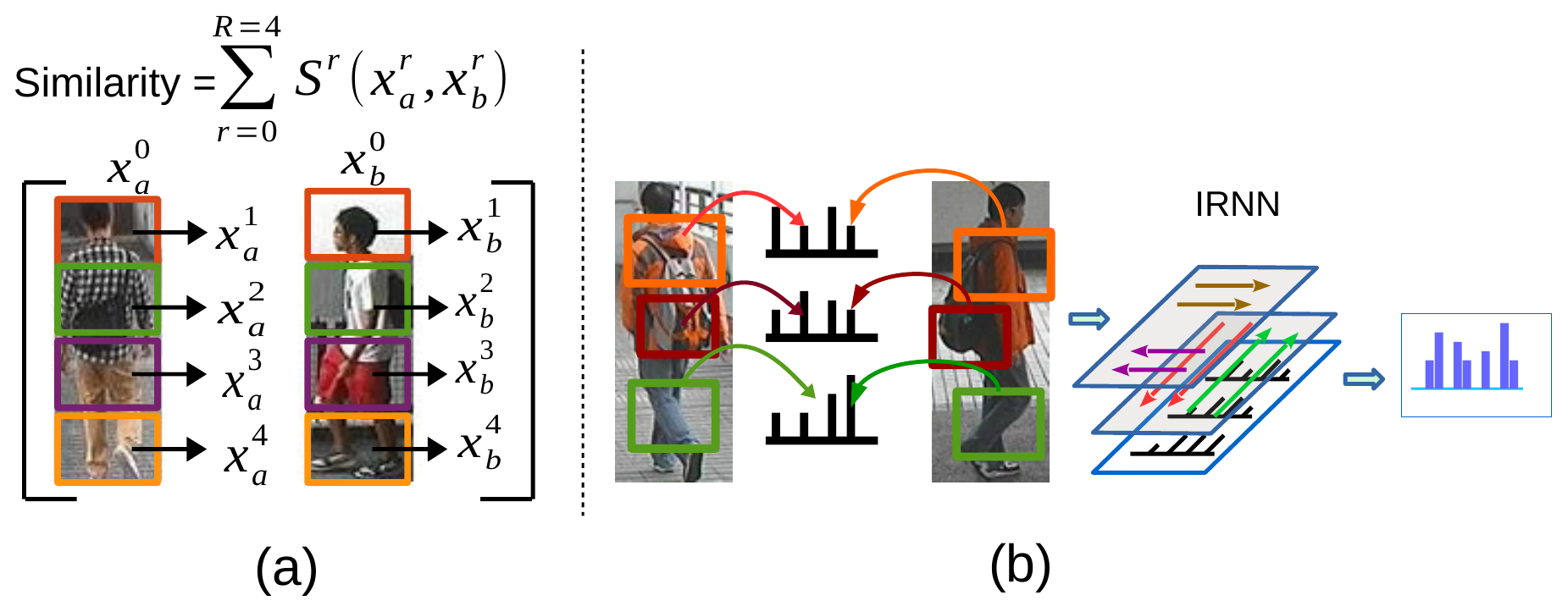}
\caption{Left: Two-stage similarity measurement divides images into regions ($r$=0 indicates the whole image), and extracts features to compute visual differences in spatial correspondences ($S^r(x_a^r,x_b^r$)). Right: Our approach learns flexible representations from local common regions and perform spatial manipulation in an end-to-end manner.}
\label{fig:motivation}
\end{figure}

Person re-id refers to matching pedestrians observed from disjoint camera views based on visual appearance. It has been attracting great attention due to its significance in visual recognition and surveillance. The major challenge in person re-id lies in the uncontrolled spatial misalignment between images due to severe camera view changes or human-pose variations. Following that, persons may resemble each other, and different identities can only be distinguished by subtle difference in the body parts and small outfit elements (\eg backpack, handbag). To this end, person re-id has benefited a lot from matching distinctive parts of persons on patch-level matching \cite{Zhao2013SalMatch,MidLevelFilter,GenerativeSaliency,eSDC} or local region aggregation \cite{Farenzena2010Person,MatchTemplate,LocalMetric}, which can address the spatial misalignment to some extent. However, these methods essentially perform a two-stage mechanism where handcrafted features are first extracted from discovered clusters of patches, and then spatial constraint is enforced to ensure spatial relationship in the matching process (See Fig.\ref{fig:motivation}). Despite their gain in performance, one salient drawback is \emph{the independence of feature extraction and localization} because feature description and spatial manipulation are individually pre-defined.

\begin{figure*}[t]
\centering
\includegraphics[height=5.5cm]{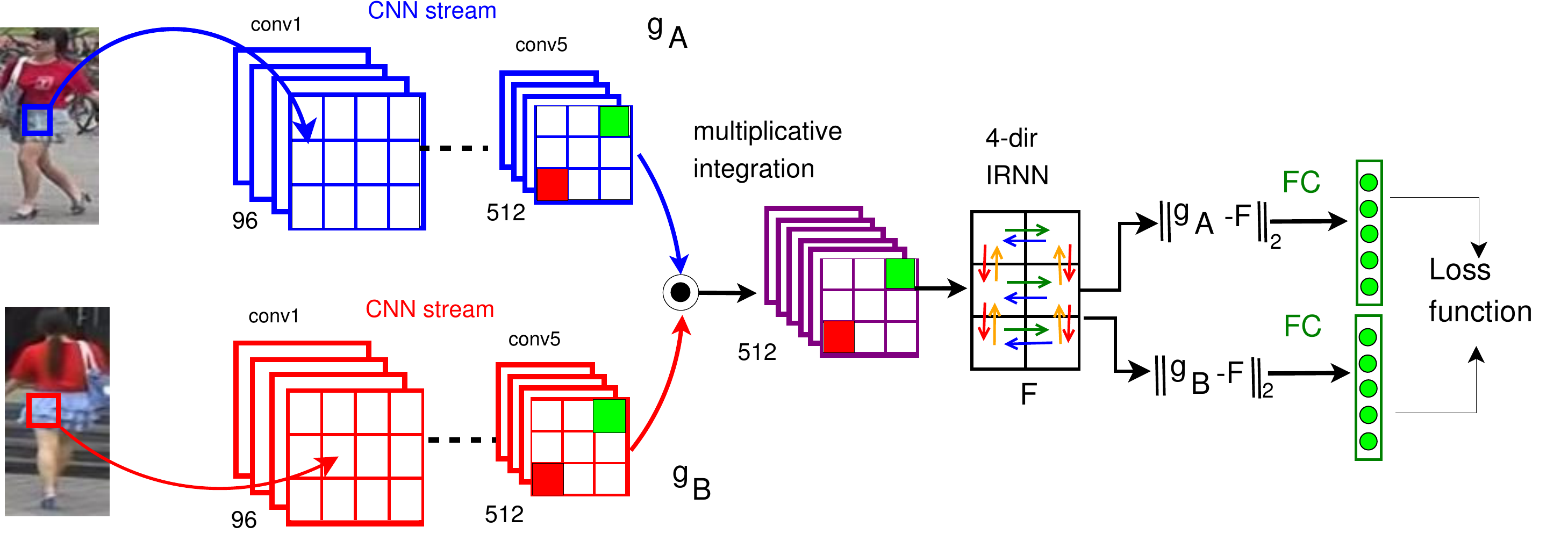}
\caption{The architecture of deep spatially multiplicative integration networks for person re-identification. The low-layers of the model are composed of two-stream CNNs whose output of last convolution are combined by multiplicative integration gate at every location. The upper-layers correspond to stacked four-directional recurrent layers to capture spatial relationship w.r.t the whole image through lateral connections. The resulting joint features can be used for similarity measurement by addressing cross-view misalignment.}
\label{fig:multiplicative}
\end{figure*}

Recently, Convolutonal Neural Network (CNN) models \cite{FPNN,JointRe-id,GatedCNN,LCAPR,Multi-channel-part} are proposed for person re-id, and most of the frameworks are designed in a Siamese fashion that learns an embedding where similar pairs (\ie images belonging to the same identity) are close to each other whilst dissimilar pairs (\ie images belonging to different identities) are separated by a distance. The striking success of these deep learning methods can be attributed to the development of learning features from local regions in patch neighborhood. However, one limitation is they extract fixed representations for each image without knowledge of the paired images that may contain distinct common local patterns to distinguish positive pairs from negative pairs. For example, Fig.\ref{fig:motivation} (b) shows that the patches corresponding to the ``bag'' (in red box) and the ``jacket'' (in orange box) are very helpful to identify the person of interest in a different camera view. On the other hand, the \emph{crucial spatial dependencies} within convolutional activations are not exploited to increase the matching confidence. For instance, as shown in Fig.\ref{fig:motivation} (a), the regions containing the heads and upper bodies of persons should be compared in their respective correspondence. To these ends, in order to enhance matching confidence level, \emph{an ideal network should be capable of capturing and propagating the local patterns while enforcing spatial dependencies to exploit complex local and global contextual information}.

\paragraph{Our Approach} In this paper, we present a novel deep architecture with an integration gating function to extract common local patterns for an image pair to increase the similarity for positive pairs. The whole architecture is illustrated in Fig.\ref{fig:multiplicative}. Similar to Siamese CNNs \cite{GatedCNN}, we instantiate two identical VGG-Net \cite{VGG} with shared parameters and pre-trained on ImageNet, whose outputs from the last convolution are integrated via multiplicative way using the Hadamard product (element-wise multiplication) at each location of their convolution activations. The effect of multiplication naturally results in a gating type structure in which two CNN stream features are the gates of each other by reconciling their  pairwise correlations to enhance common local patterns. However, the inputs come from different modalities (disjoint camera views) with different visual statistics \footnote{In practice, the complex configurations are the combinations of view points, poses, lightings and photometric settings, and thus pedestrian images are multi-modal.}, and thus making it difficult to learn feature transformations for a pair of subregions. To this end, we propose to embed two inputs using two linear mappings, followed by Hadamard product to learn joint representations in a multiplicative way. This will promote the common local subregions along the higher layers so that the network propagates more relevant features through higher layers of the deep nets. Since the gradients with respect to each input is directly dependent on the other input in their Hadamard product, the gating mechanism alters the gradient properties by boosting the joint embedding and the back propagated gradients corresponding to the promoted local similarities. This can encourage the lower and middle layers to learn filters to extract locally similar patterns that can effectively discriminate positive pairs from negatives.

To incorporate spatial relationship into feature learning, stacked four directional recurrent neural networks (RNNs) \cite{IRNN} are employed to convert the temporal dependency learning into spatial domains. One may easily add a spatial pyramid pooling (SPP) layer \cite{SPP-Net} on top of convolution layers to aggregate local features by partitioning images from finer to coarser levels. Unfortunately, SPP still exploits local inputs due to the local receptive fields rather than the contextual information of the whole image. An alternative way is multi-scale orderless pooling \cite{Multi-scale-pooling}, which extracts CNN activations for local patches at scale levels and performs orderless VLAD pooling \cite{VLAD}. However, it cannot achieve the global contextual coherence and spatial consistency over critical patches. In contrast, we propose to apply recurrent connections to render learned joint features not only spatially-correlated but also robust against spatial transformations. Our approach does not require any patch-level correspondence annotation between image pairs as it directly integrates mid-level CNN features by joint embeddings. The convolutions, multiplicative gating function, spatially recurrent layer are end-to-end trainable for person re-id by back-propagation.

The proposed architecture is inspired by bilinear CNN \cite{BilinearCNNs} whereas our model embeds spatial recurrence coupled with a fundamentally different way to capture both local and global spatial information, rather than orderless pooling on the location of features alike the bilinear CNN \cite{BilinearCNNs}. We also remark that a standard multiplicative integration network \cite{MIRNN} is not applicable to address the promotion of pairwise correlations between common local patterns from cross-view pedestrian images. This is mainly because fusing two information flows using Hadamard product directly \cite{MIRNN} is not hypothesized to deal with the modality discrepancy in which two input images from disjoint camera views exhibit different visual statistics. This may result in an interference in gradient computation with respect to each input, which is dependent on the other input in Hadamard product of the two inputs. To this end, we propose a novel integration gating function which is designed using two linear mappings for embedding two convolutional activations, followed by Hadamard product to learn joint representations in a multiplicative way, and a linear mapping to project the joint representations into an output vector. This gating function is appealing by providing the subsequent four directional RNNs \cite{IRNN} with better generalization and easier optimization. Thus, the proposed approach is more advantageous by localizing and learning common features from critical patches of identities, which can discriminate persons and align local regions in displacement.

\paragraph{Contributions} The contributions of our work are four-fold:
\begin{itemize}
\item We present an end-to-end deep network that is able to stress common local patterns against cross view changes, and thus improving matching confidence in person re-id.
\item We propose a multiplicative integration gating function to embed two stream convolutional features into joint representations, while show that the resulting features processed by spatially recurrent pooling deliver better results than alternative spatial dependency modeling methods including global average pooling, additional convolution and SPP \cite{SPP-Net}.
\item The proposed integration gate with Hadamard product allows cross-view feature alignment and facilitate end-to-end training without introducing extra parameters.
\item Our approach is demonstrated to achieve state-of-the-art results on VIPeR, CUHK03 and Market 1501 benchmark datasets.
\end{itemize}

\section{Related Work}\label{sec:related}

\subsection{Person Re-identification}

Many person re-identification methods focus on improving feature design against severe visual appearance changes \cite{Gray2008Viewpoint,Wang2007Shape,EMDBody,LinYang2017arxiv} or seeking proper metrics to measure the cross-view appearance similarity \cite{Xiong2014Person,Kostinger2012Large,GenericMetric,Pedagadi2013Local,YangLin15SIGIR,LinYang13MM,YangLin15MM,YangIJCAI2016,Multi-taskDistance,LADF}. Since these methods do not effectively address the spatial misalignment among patch matching, their recognition results are still under-performed.

To combat the spatial misalignment, some patch-level matching methods with attention to spatial layout are proposed \cite{GenerativeSaliency,MatchTemplate,Zhao2013SalMatch,eSDC,Correspondence,Farenzena2010Person} which segment images into patches and perform patch-level matching with spatial relations. Methods in \cite{MatchTemplate,Cheng2011Custom,YangLin14PAKDD,YangInf2013,Farenzena2010Person} separate images into semantic parts (\eg head, torso and legs), and measure similarities between the corresponding semantic parts. However, these methods assumes the presence of the silhouette of the individual and accuracy of body parser, rendering them not applicable when body segmentations are not reliable. Moreover, they are still suffering mismatching between distant patches.
To avoid the dependency on body segments and reduce patch-wise mismatching, saliency-based approaches \cite{Zhao2013SalMatch,eSDC} are developed to estimate the saliency distribution relationship and control path-wise matching process.
Some metric learning approaches \cite{SimilaritySpatial,LocalMetric,LADF,YangLinTIP15,YangLinTNNLS17,YangCIKM2013,LinYang2017,LOMOMetric} make attempts to extract low-level features from local regions and perform local matching within each subregions. They aim to learn local similarities and global similarity, which can be leveraged into an unified framework. Despite their effectiveness in local similarity measurement with some spatial constraints, they have limitations in the scenarios where corresponding local regions are roughly associated.
In essence, the above methods are developed on a two-stage scheme in which feature extraction and spatial arrangement are performed separately.

With the resurgence of Convolutional Neural Networks (CNNs) in a variety of tasks such as image classification \cite{VGG,AlexNet,Xnor-net,YangLinTIP17} and frequency domain \cite{CNNpack}, a number of end-to-end deep Siamese CNN architectures \cite{FPNN,JointRe-id,PersonNet} are proposed for person re-id with the objective of projecting the images of similar pairs to be closer to each other while those of dissimilar pairs to be distant from each other. However, current networks extract fixed representations for each image without consideration on transformed local patterns which are crucial to discriminate positive pairs from negatives. In contrast, we present a model to learn flexible representations from detected common local patterns which are robust against cross-view transformations. It enables automatic interactions between common part localization, feature extraction, and similarity estimation. S-CNN \cite{GatedCNN} has some sharing with us in emphasizing finer local patterns across pairs of images, and thus flexible representations can be produced for the same image according to the images they are paired with. However, their matching gate is to compare the local feature similarities of input pairs from the mid-level, which is unable to mediate the pairwise correlations to seek joint representations effectively. Moreover, S-CNN \cite{GatedCNN} manually partitions images into horizontal stripes. This renders S-CNN unable to address spatial misalignment. In contrast, our model introduces multiplicative integration gating mechanism to learn joint representations attentively from common local patterns while subject to spatial recurrence to effectively address spatial misalignment.

\subsection{Two-stream Models}

``Two-stream" architectures have been used to analyze videos where one network models the temporal aspect, while the other network models the spatial dimensions \cite{Two-stream}. Bilinear models \cite{Bilinear} can model two-factor variations and provide richer representations than linear models. To exploit this advantage, fully-connected layers in neural networks can be replaced with bilinear pooling which yeilds the outer product of two vectors. It allows all pairwise interactions among given features. Recently, the model of Bilinear CNNs \cite{BilinearCNNs} is an application of this technique in fine-grained visual recognition that generalizes orderless texture descriptors such as VLAD \cite{VLAD}, Fisher vector and cross-layer pooling \cite{Cross-layer}. However, bilinear representations tend to be high-dimensional, limiting the applicability to computationally complex tasks. To combing information flows from two different sources, multiplicative integration can be viewed as a general way by using Hadamard product on two input sources \cite{MIRNN}. Our model is inspired by multiplicative integration while we introduce a joint embedding into the integration gating function which is capable of learning locally common patterns against cross-view changes. Meanwhile spatial dependencies are preserved into feature learning.

\section{Deep Spatially Multiplicative Integration Networks}\label{sec:approach}

\begin{figure}[t]
\centering
\includegraphics[height=4cm]{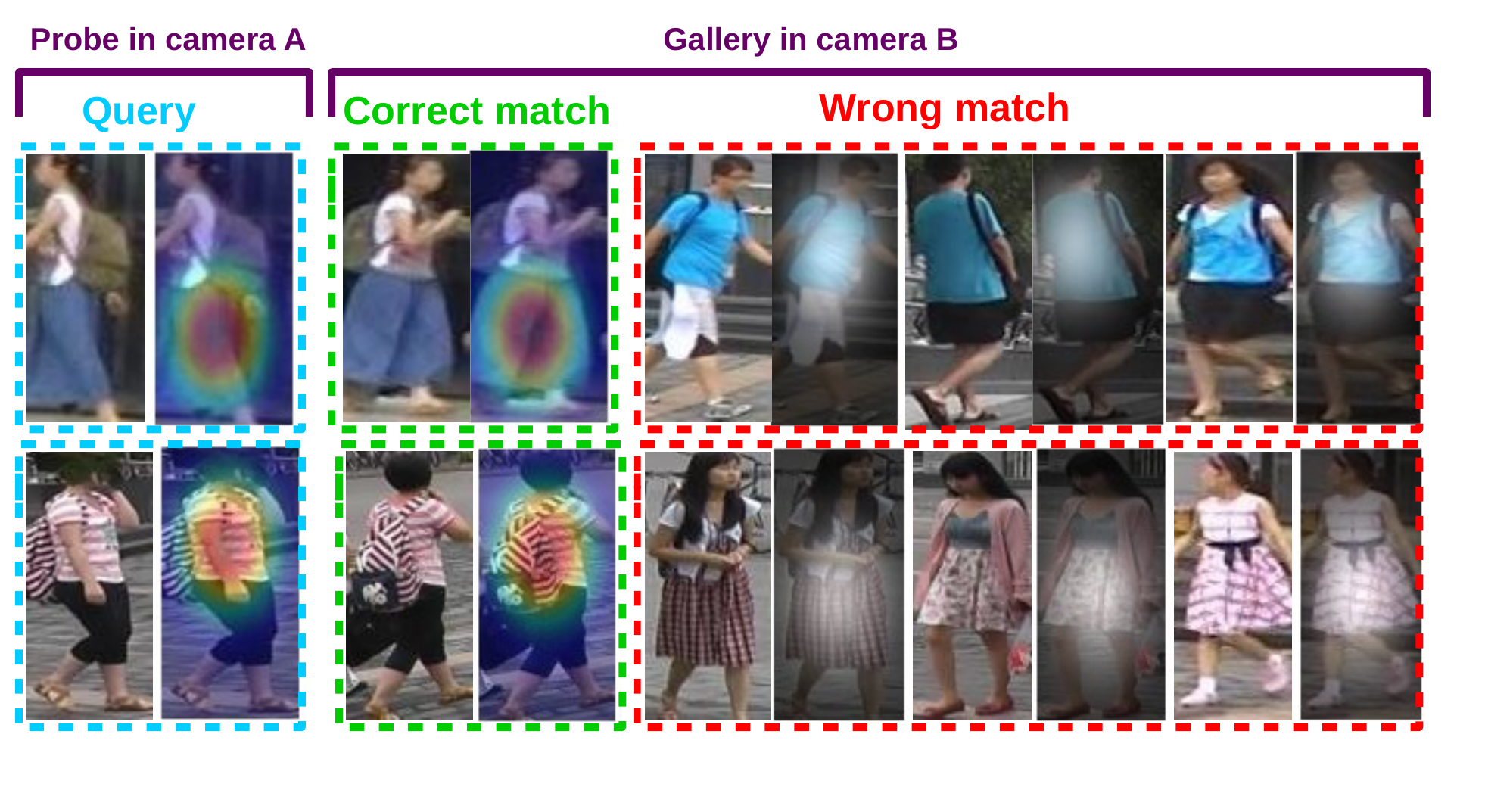}
\caption{Spatial attention effect to extract relevant features from local regions to perform matching across cameras.}
\label{fig:attention}
\end{figure}

In this section, we formulate the deep spatially multiplicative integration networks to learn locally joint representations for person re-identification.
Specifically, it can be formulated as a quadruple $\mathbb{M}=([\boldsymbol g_A, \boldsymbol g_B], \mathcal{B}, \mathcal{P}, \mathcal{L})$, where $\boldsymbol g_A$ and $\boldsymbol g_B$ are two non-linear encoders in regards to a pair of images, $\mathcal{B}$ is the multiplicative integration block, $\mathcal{P}$ is spatial pooling function, and $\mathcal{L}$ is the loss function. The overall framework
is illustrated in Fig.\ref{fig:multiplicative}, where
given the input in pairs, our model starts from two-stream convolutions (section \ref{ssec:convnet}) to localize regions and extract features, which are integrated by the Hadamard product in the multiplicative integration block at each convolution position (section \ref{ssec:MI}). The resulting features are fed into spatial recurrent pooling to propagate information through lateral connections and equip features with spatial dependencies (section \ref{ssec:spatial-recurrent}). For the similarity function, we employ the cosine similarity function and binomial deviance loss function \cite{DeepReID} (section \ref{ssec:end-end}). The whole network is end-to-end trainable and learned features are boosted to localize common distinct regions for person re-id (see Fig.\ref{fig:attention}). In what follows, we will present each component of the framework.

\subsection{Two-stream Convolutional Neural Networks}\label{ssec:convnet}

We consider CNNs to extract features from inputs in pairs. Specifically, $\boldsymbol g_A$ and $\boldsymbol g_B$ are instantiated with the VGG-Net \cite{VGG}. We pre-train the VGG-Net on the ImageNet dataset \cite{AlexNet}, and truncate at the convolutional layer including non-linearities as the feature functions. The advantage is that the resulting CNNs can process images of an arbitrary size in a single feed-forward propagation and generate outputs indexed by the location in the image and feature channels. The VGG-Net is characterized by the increased depth with very small convolution filters, which are effective on classification and localization tasks. For notational simplicity, we refer to the last convolutional layer of CNNs \ie the $5^{th}$ convolutional layer, $conv_5=\boldsymbol g_A(I)$, $conv_5'=\boldsymbol g_B(\bar{I})$ for the the image pair $[I,\bar{I}]$ as input and the activations of the last convolution as the output. This is mainly because CNNs extract low-level features at the bottom layers and learn more abstract concepts such as the parts or more complicated texture patterns at the mid-level (\ie $conv_5$). These mid-level features are more informative than higher-level features \cite{GatedCNN} and contain finer details that are crucial to increase the similarity for positive pairs. Hence, we propose a multiplicative integration gating function to attend extracted local patterns and learn flexible joint representations for an image pair.

\subsection{Multiplicative Integration Gating}\label{ssec:MI}

To integrate two CNN information flows $\boldsymbol g_A$ and $\boldsymbol g_B$, we propose a fusion design in  a form of multiplicative integration. Given the activations from the previous convolutional block $\boldsymbol g_A \in \mathbb{R}^{K\times K\times D}$ and $\boldsymbol g_B\in \mathbb{R}^{K\times K\times D}$, where $K \times K$ denote the shape of the last convolution and $D$ is feature depth, we propose to use the Hadamard product $\odot$ to fuse $g_A$ and $g_B$:
\begin{equation}\label{eq:MI}\small
\boldsymbol F(i,j,:)=\boldsymbol P^T \left( (\boldsymbol U^T \boldsymbol g_A(i,j,:)+\boldsymbol b_A)\odot (\boldsymbol V^T \boldsymbol g_B(i,j,:) +\boldsymbol b_B)\right) + \boldsymbol b,
\end{equation}
where $0 \leq i,j \leq K$, $\boldsymbol F \in \mathbb{R}^{K\times K\times D}$ denote the integrated features, and each vector of $\boldsymbol F(i,j,:)$ is determined by the two linear mappings $\boldsymbol U \in \mathbb{R}^{D\times d}$, $\boldsymbol V \in \mathbb{R}^{D\times d}$ for embedding two input vectors $\boldsymbol g_A(i,j,:)$ and $\boldsymbol g_B(i,j,:)$, by the Hadamard product $\odot$ (element-wise multiplication). $d$ is a hyper-parameter to decide the dimension of joint embedding. $\boldsymbol P \in \mathbb{R}^{d \times D}$ denotes the linear mapping with a bias $\boldsymbol b \in \mathbb{R}^D$ to project the joint representations into an output vector. $\boldsymbol b_A \in \mathbb{R}^d$ and $\boldsymbol b_B \in \mathbb{R}^d$ are bias vectors for their respective linear projections $\boldsymbol U$ and $\boldsymbol V$.

Applying non-linear activation functions may help to increase representative capacity of the model. We apply non-linear activation right after linear mappings for input vectors:
\begin{equation}\label{eq:non-linear}
\boldsymbol F(i,j,:)=\boldsymbol P^T \left( \sigma(\boldsymbol U^T \boldsymbol g_A(i,j,:))\odot \sigma(\boldsymbol V^T \boldsymbol g_B(i,j,:) )\right) + \boldsymbol b,
\end{equation}
where $\sigma$ denotes the sigmoid activation function, which maps real values into a finite interval [0,1]. The biases $\boldsymbol b_A$ and $\boldsymbol b_B$ are omitted for brevity. The effect of this multiplication naturally results in a gating type structure, in which $\boldsymbol U^T \boldsymbol g_A(\cdot,\cdot,\cdot)$ and $\boldsymbol U^T \boldsymbol g_B(\cdot,\cdot,\cdot)$ are the gates of each other. More specifically, the embedding of input features ($\boldsymbol g_A$) can be influenced by its positive match ($\boldsymbol g_B$), through which common local patterns are amplified into the joint embeddings/representations. The proposed gating unit in Eq.\eqref{eq:non-linear} aggregates the information flow from $\boldsymbol g_A$ and $\boldsymbol g_B$ while allowing them to be conditioned by considering their pairwise interactions. Moreover, this integration introduces no extra parameters since the bias vectors $\boldsymbol b_A$ and $\boldsymbol b_B$ are negligible compared to the total number of parameters. Also, it offers the advantage by providing better generalization and easier optimization wherein the gradient properties are changed due to the gating effect and most of hidden units are non-saturated.

\begin{figure}[t]
\centering
\includegraphics[height=3cm]{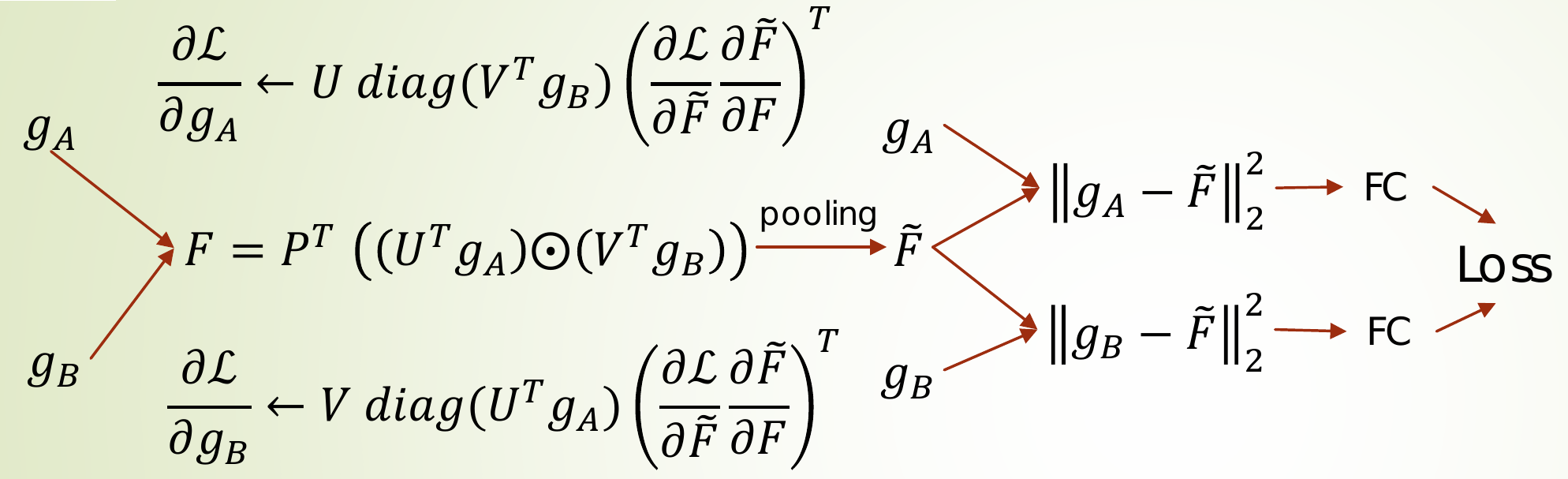}
\caption{Computing gradients in the multiplicative integration gating.}
\label{fig:gradient}
\end{figure}

\paragraph{Gradient properties}
The multiplicative integration gating has different gradient properties by regulating the gradient flow conditioned on each other. Let $\frac{\partial \mathcal{L}}{\partial \boldsymbol F}$ be the gradient of loss function $\mathcal{L}$ w.r.t $\boldsymbol F$, then by the chain rule of gradients we have:
\begin{equation}
\frac{\partial \mathcal{L}}{\partial \boldsymbol g_A}= \boldsymbol U \mbox{diag}(\boldsymbol V^T \boldsymbol g_B) \left(\frac{\partial \mathcal{L}}{\partial \boldsymbol F}\right)^T, \frac{\partial \mathcal{L}}{\partial \boldsymbol g_B}=\boldsymbol V \mbox{diag}(\boldsymbol U^T \boldsymbol g_B)  \left(\frac{\partial \mathcal{L}}{\partial \boldsymbol F}\right)^T.
\end{equation}
By looking at the gradients, we see that the matrix $\boldsymbol V$ and the paired input $\boldsymbol g_B$ are directly involved in the gradient computation by gating the matrix $\boldsymbol U$, hence more capable of altering the updates of the learning nets. Fig.\ref{fig:gradient} illustrates the gradient scheme. While the outputs from two CNN streams are combined at each spatial location in a multiplicative way, the spatial context is not preserved. To this end, a spatially recurrent pooling $\mathcal{P}$ over all locations is subsequently performed. The resulting features are flattened and fed into a loss function $\mathcal{L}$ to determine the matching measure.

\subsection{Spatially Recurrent Pooling with IRNNs}\label{ssec:spatial-recurrent}

\begin{figure*}[t]
\centering
\includegraphics[height=5cm]{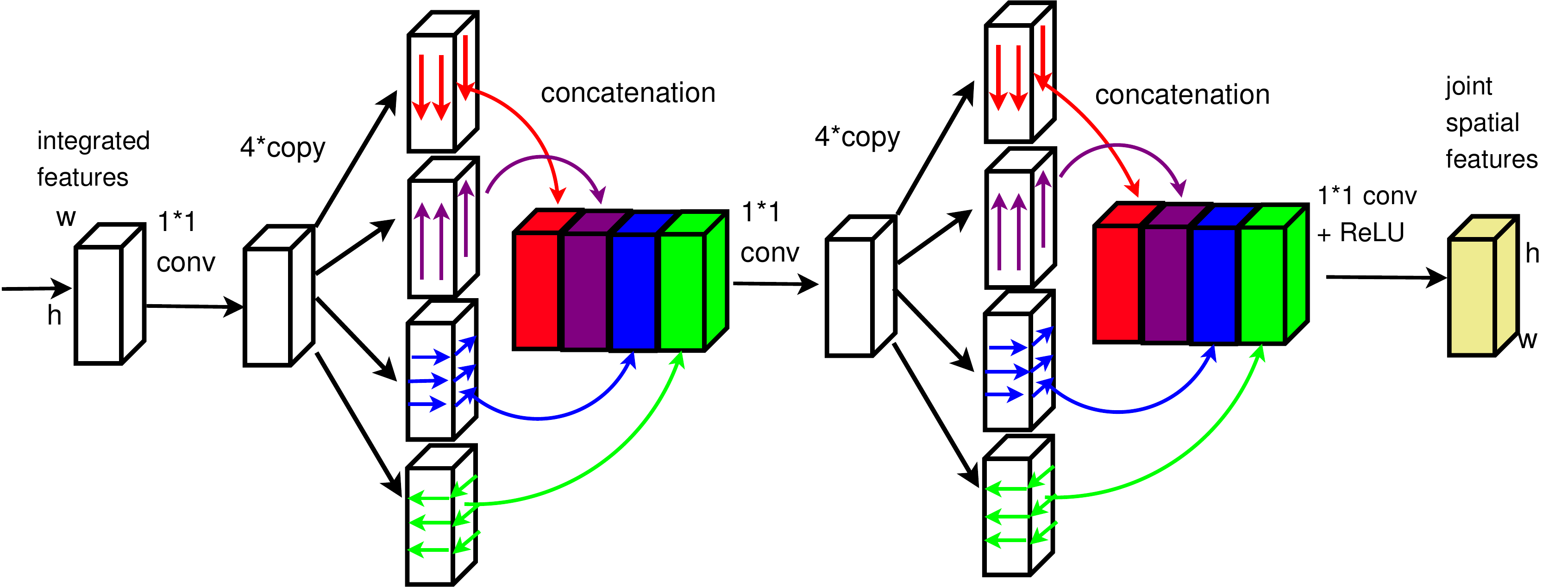}
\caption{The architecture of four-directional IRNN. The spatial RNNs are instantiated using ``IRNN" units that are composed of RNNs with ReLU recurrent transitions, and initialized to the identity matrix. All transitions from/to the hidden states are computed with $1\times 1$ convolutions, which enables more efficient computation in recurrence. The final spatially recurrent pooled features are outputs of two-layer IRNNs, where the spatial resolutions remain the same as bilinear features.}
\label{fig:iRNN}
\end{figure*}

Our architecture for incorporating spatial dependency into feature learning is shown in Fig.\ref{fig:iRNN}. This action explicitly allows the spatial manipulation of feature maps within the network, and can effectively address the spatial alignment issue in matching persons in cross-view setting.
We employ four RNNs that sweep over the entire feature map in four different directions \cite{ReNet}: bottom to  top, top to bottom, left to right, and right to left.
The recurrent layers ensure that each feature activation in its output is an activation at a spatially specific location with respect to the whole image.

On top of the integrated features $\boldsymbol F$, we place RNNs that move \emph{laterally} across the activations, and produce an output $\widetilde{\boldsymbol F}$ the same size as $\boldsymbol F$. Thus, the temporal dependency learning in RNNs is converted to spatial domain. The recurrent neural networks can be implemented using several forms: long short-term memory (LSTM) \cite{lstm1997}, gated recurrent units (GRU) \cite{GRU2014}, and plain \textit{tanh} recurrent neural networks. More recently, Le \etal \cite{IRNN} show that RNNs composed of rectified linear units (ReLU) are easily to train and are supreme in modeling long-range dependencies if the recurrent weight matrix is initialized to the identity matrix. A ReLU RNN initialized in this way is named as ``IRNN" \cite{IRNN} \footnote{In initialization, gradients are propagated back-towards with full strength.}, and it performs almost as well as an LSTM for a real-world language modeling task. In this paper, we adopt this architecture on account of its simple implementation and faster computation than LSTMs and GRUs.

In our model, four independent IRNNs that move in four directions are applied upon integrated features and IRNNs can be efficiently computed by splitting the internal IRNN computations into separate logical layers \cite{IONet}. As shown in Fig. \ref{fig:iRNN}, the input-to-hidden transition is a $1\times1$ convolution, which can be shared across different directions. The bias term is shared in the same way and merged into the convolution. Sharing transitions in this way can accelerate the IRNN computation while without dropping accuracy \cite{IONet}. As a result, the IRNN layer corresponds to apply recurrent matrix and the nonlinearity at each step, from which the output is computed by concatenating the hidden states from the four directional IRNNs at each spatial location. Each IRNN takes the output of $1\times 1$ convolution, and updates its hidden states via:
\begin{equation}\label{eq:hidden}
\widetilde{F}_{i,j}^{\mbox{dir}} \leftarrow \max\left( W_{hh}^{\mbox{dir}} \widetilde{F}_{i,j-1}^{\mbox{dir}}+ \widetilde{F}_{i,j}^{\mbox{dir}}, 0\right),
\end{equation}
where $\mbox{dir}$ indicates one of the four directions that moves to $\mbox{dir}\in \{left, right, up, down\}$. In Eq.\eqref{eq:hidden}, the input-to-hidden transition is not presented because it has been computed as part of the $1\times1$ convolution, and then copied in-place to each hidden recurrent layer. In our implementation, the number of hidden units is 512.

\begin{figure}[t]
\centering
\includegraphics[height=4cm]{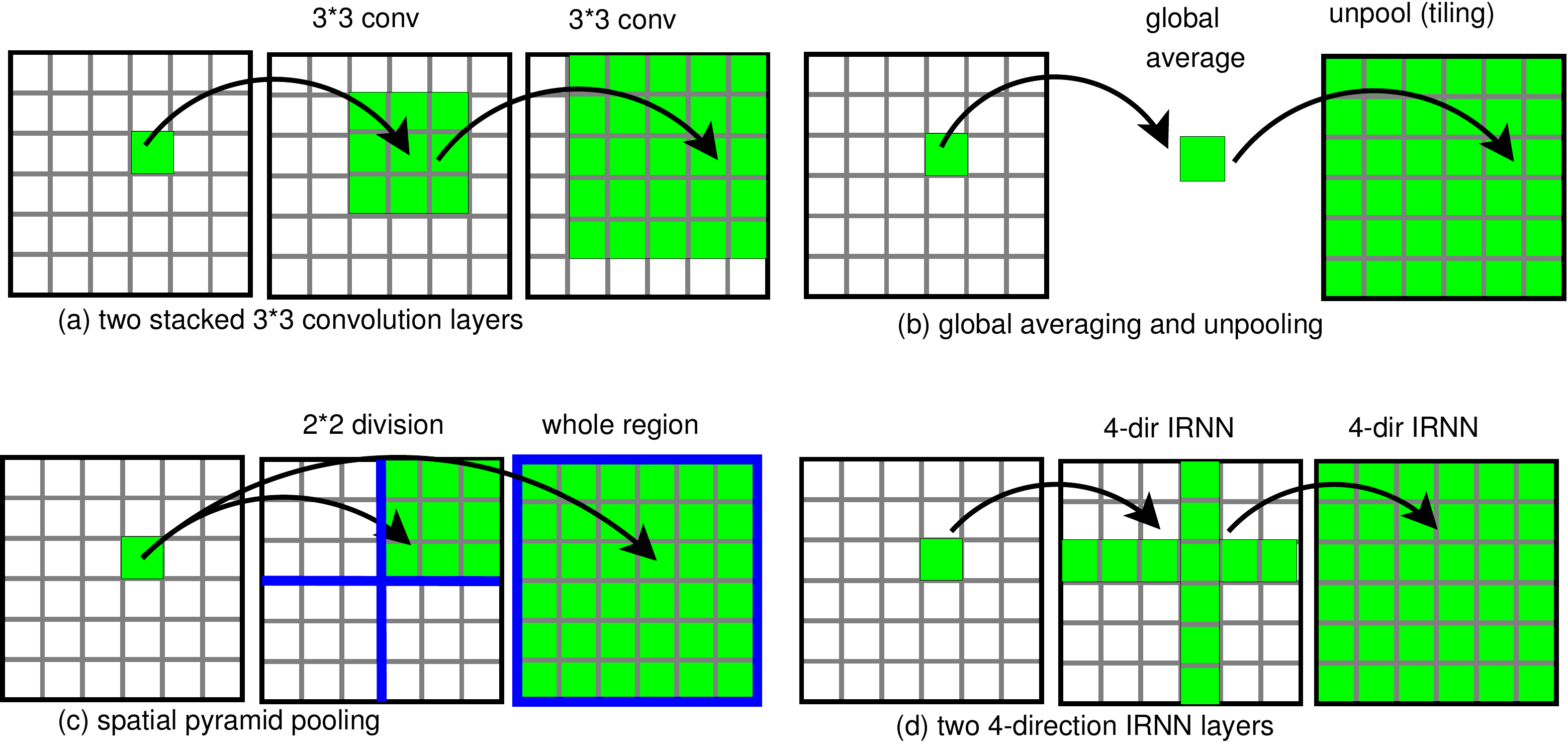}
\caption{Incorporating context with different respective fields. (a) $2\times$ stacked convolutions on top of bilinear features allow a cell in input to affect a $5\times 5$ window in output. (b) In global average pooling, each cell in the output depends on the entire input with the same value repeated. (c) In spatial pyramid pooling, each cell in the input is put through by a multi-level pooling. (d) In 4-dir IRNN, each cell in the output depends on the entire input with spatially varied value.}
\label{fig:spatial-pooling}
\end{figure}

\paragraph{Comparison with alternative contextual models}
It is known that RNNs are not the exclusive way of incorporating contextual information. Common ways of embedding contextual and spatial priors include global average pooling \cite{ParseNet}, additional convolution layers, and spatial pyramid pooling \cite{PyramidKernel2005,SPM2006}.
As shown in Fig. \ref{fig:spatial-pooling} (b),  global pooling provides information about the entire image and one could apply a global average and unpool (tile or repeat spatially) back to the original feature map as conducted in ParseNet \cite{ParseNet}.
One could also simply add additional convolution layers on top of integrated features and then pool out of the top convolution layer. For example, in Fig. \ref{fig:spatial-pooling} (a), stacked $3\times 3$ convolutions can add two cells worth of context.
A spatial pyramid pooling layer \cite{SPP-Net} can be added on top of $\boldsymbol F$ where spatial information is maintained by max-pooling in local spatial bins (Fig.\ref{fig:spatial-pooling} (c)).

\begin{figure}[t]
\centering
\includegraphics[height=3cm]{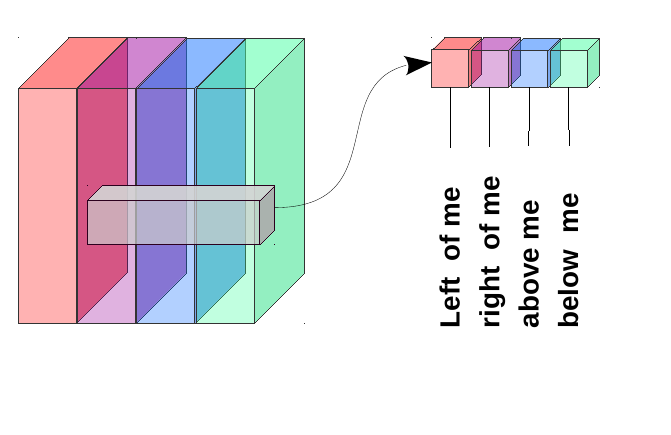}
\caption{The output of the first IRNN layer. Each cell in the output summarizes the features with respect to the left/right/top/down.}
\label{fig:first_output}
\end{figure}

Compared with alternatives, the two stacked 4-dir IRNN layers have fewer parameters and can propagate information through lateral connections that span across the whole image. Our empirical studies on test set in person re-id show that stacked 4-dir IRRNs are able to achieve better performance than alternatives (in Section \ref{sec:exp} Table \ref{tab:com_cmc}). In fact, after the first 4-dir IRRN, a feature map is produced that summarizes nearby objects at every position in the image, that is, the first IRRN can create a summary of the features to the left/right/top/down of every cell, as illustrated in Fig.\ref{fig:first_output}. The subsequent $1\times 1$ convolution mix these priors as a dimension reduction. The second 4-dir IRRN can ensure every cell on the output depends on every cell of the input, and producing contextual features both global and local. In this way, the features vary locally by spatial position, while each cell is a global summary of the image with respect to that specific spatial location.

\subsection{Loss Function}\label{ssec:end-end}

The last layer in the network uses a similarity function to measure the whether two images ($i,j$) belong to the same person or not given the output features learned by the deep model. Once the spatially integrated features $\widetilde{\boldsymbol F}$ are trained, we calculate the Euclidean distance between $\boldsymbol g_A$,$\widetilde{\boldsymbol F}$ and $\boldsymbol g_B$,$\widetilde{\boldsymbol F}$, respectively, that is, $||\boldsymbol g_A-\widetilde{\boldsymbol F}||_2$, $||\boldsymbol g_B-\widetilde{\boldsymbol F}||_2$. Then, fully-connect layer is added to produce their final representations which can be fed into a loss function. Specifically, we use the cosine similarity function and the binomial deviance loss function for training:
\begin{equation}\label{eq:loss}
\mathcal{L} = \sum_{i,j} W \odot \mbox{ln}(\exp ^{-\alpha (S-\beta) \odot M} + 1)
\end{equation}
where $\odot$ is element-wise multiplication, $i$ and $j$ are the number of training images, and $S=[S_{i,j}]_{n \times n}$ is the similarity matrix for image pairs ($n$ is total number of training images. $S_{i,j}=cosine(x_i,x_j)$).  $\alpha$ and $\beta$ are hyper parameters. The matrix $M$ encodes the training supervision that is defined as \[M=\left\{\begin{array}{cl}
1,& \mbox{positive pair}\\
-1,& \mbox{negative pair}\end{array}\right.\]
$W$ indicates a weight matrix that is defined as
\[W_{i,j}=\left\{\begin{array}{cl}
\frac{1}{n_1},& \mbox{positive pair}\\
\frac{1}{n_2},& \mbox{negative pair}\end{array}\right.\]
where $n_1$ and $n_2$ are the count of positive and negative pairs.

The network can be trained by back-propagation gradients of the loss function. The integrated form simplifies the gradients at the gating layer, and the recurrent layer is smooth, and continuous function. The gradients of the loss function \eqref{eq:loss} is straightforward \cite{DeepReID}, and the gradients of the layers below the multiplicative integration layer can be computed using the chain rule.

\section{Experiments}\label{sec:exp}

\subsection{Datasets}

We perform experiments on three benchmarks for person re-id: VIPeR  \cite{Gray2007Evaluating}, CUHK03 \cite{FPNN}, and Market-1501 \cite{Market1501}.
\begin{itemize}
\item The \textbf{VIPeR} dataset contains $632$ individuals taken from two cameras with arbitrary viewpoints and varying illumination conditions. The 632 person's images are randomly divided into two equal halves, one for training and the other for testing.
\item The \textbf{CUHK03} dataset includes 13,164 images of 1360 pedestrians. It is captured with six surveillance cameras. Each identity is observed by two disjoint camera views, yielding an average 4.8 images in each view. We perform experiments on manually labeled dataset with pedestrian bounding boxes. The dataset is randomly partitioned into training, validation, and test with 1160, 100, and 100 identities, respectively.
 \item The \textbf{Market-1501} data set contains 32,643 fully annotated boxes of 1501 pedestrians, making it the largest person re-id dataset to date. Each identity is captured by at most six cameras and boxes of person are obtained by running a state-of-the-art detector, the Deformable Part Model (DPM) \cite{MarketDetector}.  The dataset is randomly divided into training and testing sets, containing 750 and 751  identities, respectively.
\end{itemize}

\subsection{Implementations}

It is known that augmenting training data often leads to better generalization \cite{AlexNet}. We carry out two primary data augmentation in experiments: flipping and shifting. For flipping, we flipped each sample horizontally, which allows the model observe mirrored images of the original during training. For shifting, we shift the image by 5 pixels to the left, 5 pixels to the right and after this processing, we further shift the image by 10 pixels to the top, 10 pixels to the bottom. This two-step shifting procedure makes the model more robust to slight shifting of a person. The shifting was done with padding the borders of images. Training pairs in each batch are formed by randomly selecting 128 images from all cameras. The label for each training pair is assigned accordingly to the identity number (+1 for the same identity, -1 for different identities). The training batch is shuffled in each epoch to ensure the network can see divergent image pairs during training. Parameters in the cost function are set as $\alpha=2$, $\beta=0.5$. Even though each mini-batch that contains randomly chosen images has very small number of positive pairs, it will not impair the learning process since the binomial deviance loss is weighted in align with the number of positive/negative pairs. We deploy the dropout on recurrent layers with probability of 0.5, while each optimization run is early stopped based on validation error. We employ the VGG-Net model \cite{VGG} to extract CNN features, and the outputs of the last convolution with non-linearities are used as features with 512-dim features at each location. In our experiments, we adopt the widely used single-shot modality with Cumulative Matching Characteristic (CMC) as metric. This evaluation is performed ten times, and average CMC results are reported.

\subsection{Baselines}

The proposed method contains two novel ingredients: (1) the multiplicative integration computation that is able to attend common local regions helpful in matching, and (2) the spatially recurrence that serves to embed spatial dependencies into feature learning. To reveal how each ingredient contributes to the performance improvement, we consider the following four deep baselines:
\begin{itemize}
\item (1) CNN with fully-connected layers (FC-CNN):
The input image is resized to $224\times 224$ and mean-subtracted before propagating it through CNN. For fine-tuning, we replace the 1000-way classification layer with a $k$-way softmax layer where $k$ is the number of identity classes in each person re-id dataset. The parameters of the softmax layer are initialized randomly and the training is stopped by monitoring the validation error. The layer before softmax layer is used to extract features.
\item (2) Fisher vectors with CNN features (FV-CNN):
We construct a descriptor using FV pooling of CNN filter bank responses with 64 GMM components \cite{FilterBanks}. FV is computed on outputs of the last convolution layer of CNN.
\item (3) Fisher vectors with SIFT features (FV-SIFT):
It uses dense SIFT features \cite{MidLevelFilter} over a set of 14 dense overlapping $32\times 32$ pixels regions with a step stride of 16 pixels in both direction. The features are PCA projected before learning a GMM with 256 components.
\item (4) Bilinear CNN \cite{BilinearCNNs} with spatial pyramid pooling \cite{SPP-Net} (B-CNN+SPP): To have fair comparison, we perform a 2-level pyramid \cite{SPP-Net}: $2\times 2$ and $1\times 1$ subdivisions over the resulting bilinear features.
\item (5) Bilinear CNN \cite{BilinearCNNs} with stacked four directional RNNs \cite{IRNN} (B-CNN+IRNNs): It uses four RNNs that move four directions upon the bilinear pooling features to preserve the spatial manipulation.
\item (6) Multiplication Integration networks \cite{MIRNN} with spatial pyramid pooling \cite{SPP-Net} (MI+SPP): Two identical VGG-Net are used to extract features from images, and the last convolutions are integrated by the Hadamard product, followed by the SPP to impose spatial relationships.
\item (7) Our approach: We consider three variants varied on the initialization of the two-stream CNNs: (i) initialized by two VGG M-Nets \cite{M-Net} (Ours [VGG-M,VGG-M]); (ii) initialized by a VGG D-Net \cite{VGG} and an M-Net (Ours [VGG-D, VGG-M]); (iii) initialized by two VGG D-Nets (Ours [VGG-D, VGG-D]). The M-Net is characterized by the decreased stride and smaller receptive field of the first convolutional layer. The D-Net increased the depth with very small convolution filters, which is effective on classification and localization tasks. For both M-Net and D-Net,  The input paired images are resized to $224\times 224$ and feature are truncated at the last convolutional layer
including non-linearities as the feature functions. Two stream features are then put through integration gating function and spatially recurrent pooling. The recurrent layer has 512 hidden states for each location's encoding. Thus, the final spatially recurrent feature is of size $512\times 196$, which is comparable to FV-CNN ($512\times 128$) and FV-SIFT ($512 \times 80$). For fine-tuning, we first initialize the convolution layers using fine-tuned FC-CNN and then the entire model is fine-tuned with loss function using back-propagation.
\end{itemize}

\subsection{Comparison to Baselines}

\begin{table*}[t]
\small
\caption{CMC result at rank-R (R=1,10,20) recognition rate. }\label{tab:com_cmc}
\begin{tabular}{l|ccc|ccc|ccc}
\hline
Dataset & \multicolumn{3}{c|}{VIPeR} & \multicolumn{3}{c|}{CUHK03} & \multicolumn{3}{c}{Market-1501}\\
\cline{1-10}
Rank @ R & $R=1$ & $R= 10$ & $R=20$ & $R=1$ & $R=10$ & $R=20$ & $R=1$ & $R=10$ & $R=20$ \\
\hline
FV-SIFT & 35.89 &  68.82 & 78.94  & 49.10 & 69.32  & 77.40  & 53.35 & 82.45& 85.78\\
FC-CNN  & 32.64  & 64.33 & 70.82 & 46.14 & 62.08 & 68.11  &51.23&79.73&84.56\\
FV-CNN  & 40.59  & 75.33 & 83.52 & 50.51 & 74.62 & 80.60  &54.66&82.33&86.05\\
B-CNN + SPP &  47.59 & 83.68 & 89.27 & 56.33 & 84.55 & 91.05 &60.76& 88.02& 89.13\\
B-CNN + IRNNs & 48.27 & 84.35 & 90.46 & 58.29 & 85.71 & 92.82 & 62.33 & 89.97 & 91.05 \\
MI + SPP & 48.41 & 84.98 & 90.22 & 61.42 & 88.40 & 93.21 & 64.07 & 91.85 & 92.48 \\
\hline
Ours  (VGG-M, VGG-M) & 48.60 & 85.33  & 90.79 & 70.40  & 93.92 & 95.30 & 65.78& 93.67 & 95.08 \\
Ours  (VGG-D, VGG-M) & 48.55 & 84.75 & 90.05 & 69.22 & 93.56 & 94.37&64.96 & 93.27 & 94.57\\
Ours  (VGG-D, VGG-D) &  \color{red}$\mathbf{49.11}$  & \color{red}$\mathbf{87.66}$ & \color{red}$\mathbf{93.47}$ & \color{red} $\mathbf{73.23}$ & \color{red} $\mathbf{96.73}$ & \color{red}$\mathbf{97.52}$  & \color{red}$\mathbf{67.15}$ & \color{red}$\mathbf{95.67}$ & \color{red}$\mathbf{97.08}$\\
\hline
\end{tabular}
\end{table*}

We report the CMC values at the ranking list on three datasets attained from the proposed approach and baseline methods. The results are shown in Table \ref{tab:com_cmc},  where FV-CNN outperforms FV-SIFT and FC-CNN. One reason is that FV pools local features densely within the described regions but removing the global spatial information, hence it is more apt at describing local regions. Our model achieves the best performance consistently in all cases. In particular, our approach characterizes a notable margin in rank-1 value compared with B-CNN + SPP. For instance, on VIPeR dataset, B-CNN+SPP achieves 47.59\% on rank-1 while our method (VGG-D, VGG-D) achieves 49.11\%. On CUHK03 and Market-1501, B-CNN + SPP achieves 56.33\% and 60.76\% while our method improves the accuracy value by 17\% and 7\%, respectively.

To examine the component effectiveness contributed by multiplicative integration and IRNNs, we calculate the CMC values from two baselines: B-CNN + IRNNs and MI + SPP. The results show that B-CNN + IRRNs outperforms B-CNN + SPP in all rankings due to the effectiveness of recurrence which renders temporal dependency into spatial domains. On the other hand, MI + SPP has noticeable performance gain over B-CNN + SPP and B-CNN + IRNNs. This suggests the importance of MI to person re-id for its effectiveness in seeking finer common local patterns across views. In addition, comparative results from three variants of our model show that two identical CNN streams based on VGG-D Net are superior to alternatives using VGG-M Nets and/or VGG-D Net. The main reason is two identical CNNs with shared parameters are suitable to person re-id to extract common patterns. Thus, our method (VGG-D, VGG-D) is used as the default in all experiments.

\subsection{Comparison to State-of-the-art Approaches}

\begin{figure*}[t]
   \begin{tabular}{cc}
   \centering
     \includegraphics[width=3in,height=2.2in]{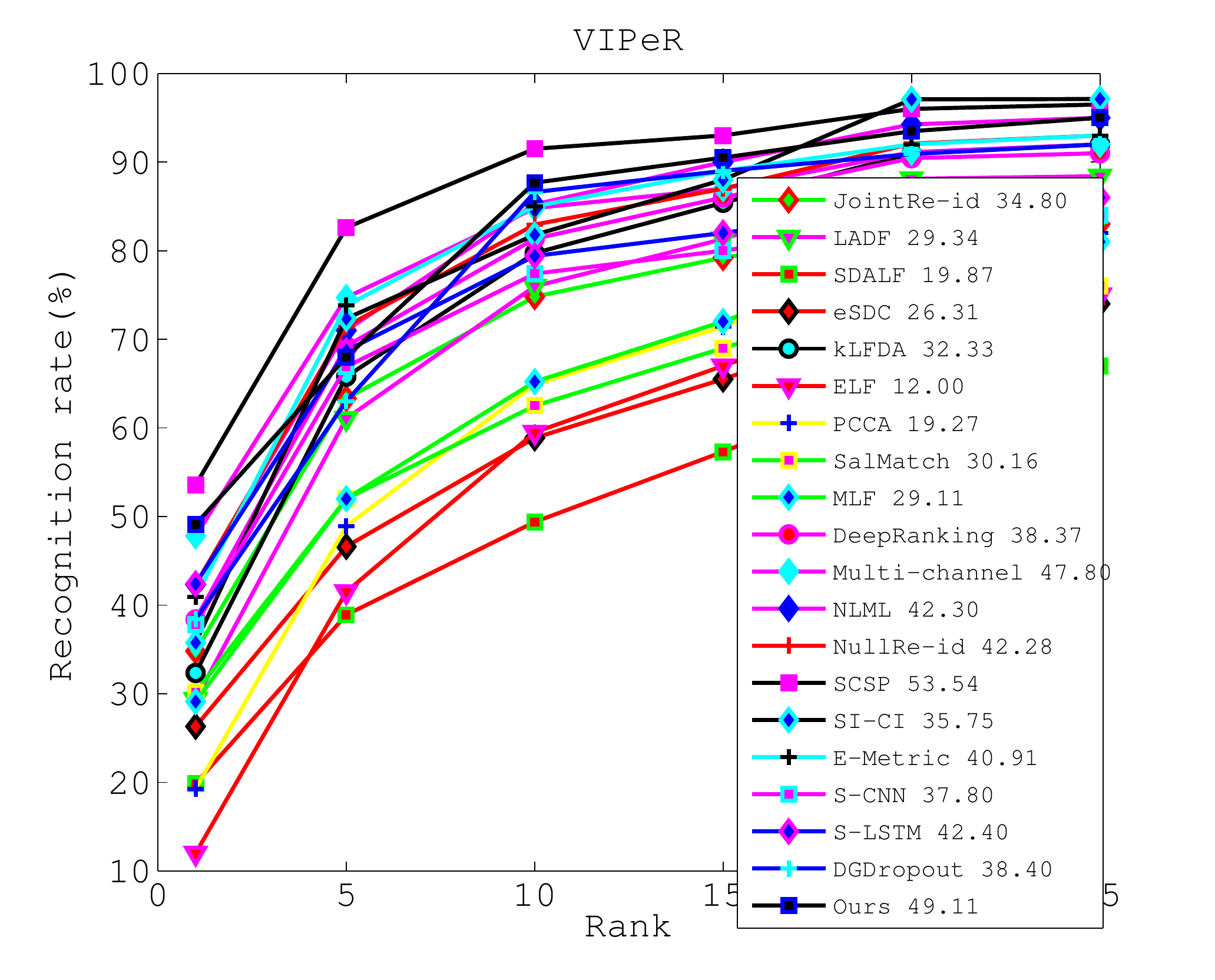}
	\includegraphics[width=3in,height=2.2in]{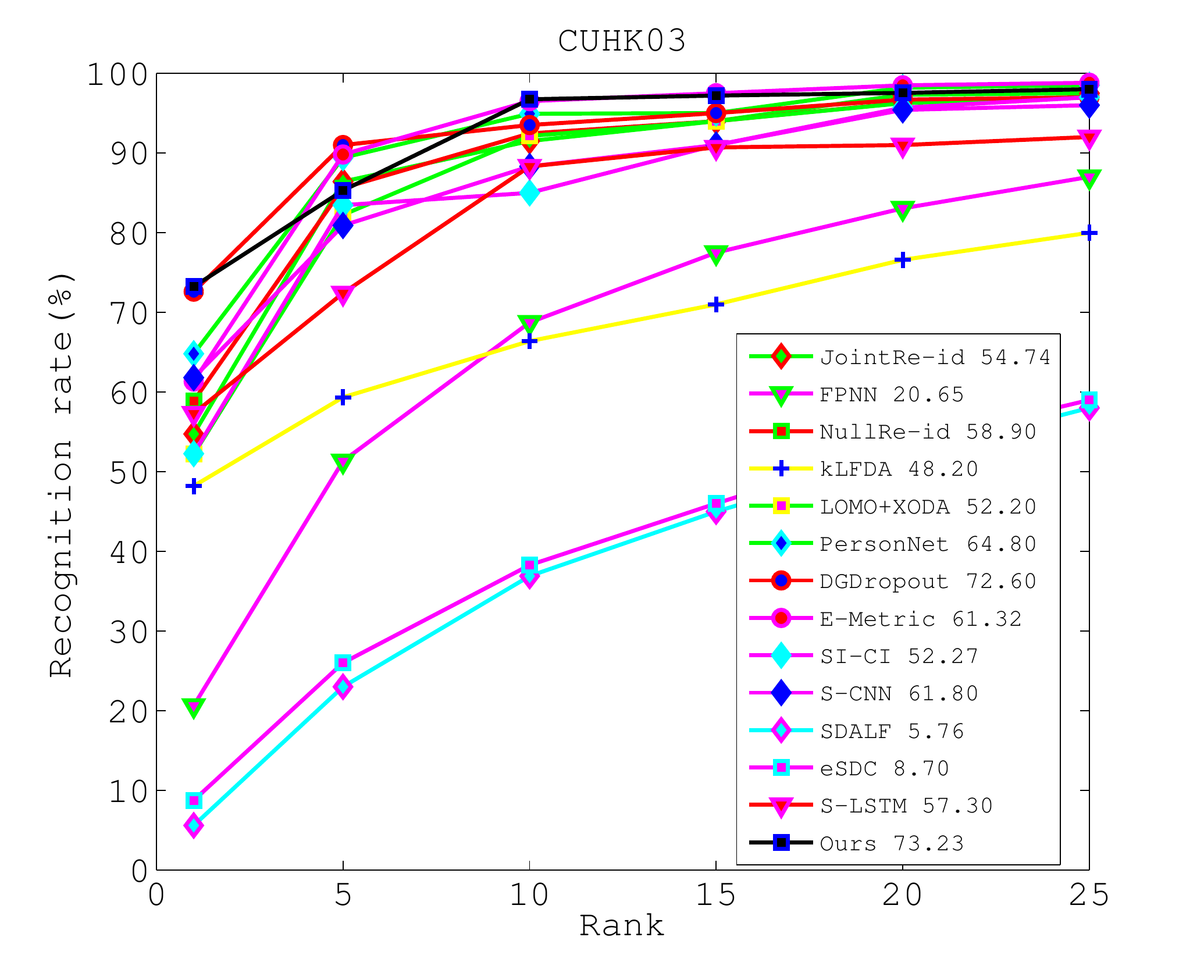}
\end{tabular}\caption{Performance comparison with state-of-the-art approaches using CMC curves on VIPeR, and CUHK03 data sets.}\label{fig:cmc}
\end{figure*}

In this experiment, we evaluate the proposed method by comparing to the state-of-the-art approaches. The CMC values on the ranking list are reported in Table \ref{tab:cmc_viper} - \ref{tab:cmc_market}, and Fig.\ref{fig:cmc}. Compared to the methods based on path-matching with additional spatial constraint, such as SDALF \cite{Farenzena2010Person}, eSDC \cite{eSDC}, SalMatch \cite{Zhao2013SalMatch}, and NLML \cite{LocalMetric}, our approach outperforms consistently by performing localization and spatial manipulation jointly. Compared to recent CNN models including JointRe-id \cite{JointRe-id}, Multi-channel \cite{Multi-channel-part}, SI-CI \cite{SI-CI}, S-CNN\cite{GatedCNN}, and S-LSTM \cite{S-LSTM}, our method has performance gain by introducing integration gating function to produce flexible representations by attending common patterns and also spatial recurrent layer to effectively address spatial misalignment. We remark that on VIPeR dataset, our method is slightly inferior to SCSP \cite{SimilaritySpatial} in which SCSP achieves 53.54\% at rank-1 accuracy while our method attains 49.11\%. This is because VIPeR is very small and does not have sufficient training samples for each identity to predict common local patterns faithfully through deep nets. Comparison results with respect to more recent state-of-the-art PIE \cite{PIE-reid} and Supervised Smoothed Manifold (SSM) \cite{SSM} suggest a number of observations. First, our method improves PIE \cite{PIE-reid} by rank-1 accuracy value 31.01, 10.83, and 1.47 for VIPeR, CUHK03, and Market-1501, respectively. The PIE \cite{PIE-reid} is a robust pedestrian descriptor which addresses the misalignment by introducing pose invariant embedding. However, they extract fixed representations from body parts of each pedestrian bounding box and fuse them into a pose invariant figure. This manual extraction is unable to learn flexible representations that account for common local patterns in the paired images.  Second, the performance improvement over SSM \cite{SSM} + GOG \cite{GOG} on CUHK03 is increased by 1.41. This gain is not dramatic due to the affinity metric calculated under the XQDA metric learning method \cite{LOMOMetric}. In Market-1501 dataset, our method combined with XQDA \cite{LOMOMetric} outperforms SSM \cite{SSM} by 3.94 at rank-1.

\begin{table*}[hbt!]
  \centering
  \caption{Rank-1,  -10, -20 recognition rate of various methods on the VIPeR data set (test person =316). }  \label{tab:cmc_viper}
  {
  \begin{tabular}{l|c|c|c}
  \hline
\hline
    Method  & $R=1$   & $R=10$  & $R=20$ \\
  \hline
   JointRe-id \cite{JointRe-id} & 34.80  & 74.79 & 82.45  \\
   LADF \cite{LADF}  & 29.34 & 75.98 & 88.10\\
   SDALF \cite{Farenzena2010Person} & 19.87  & 49.37 & 65.73\\
   eSDC \cite{eSDC} & 26.31  & 58.86 & 72.77\\
   kLFDA \cite{Xiong2014Person} & 32.33  & 79.72 & 90.95\\
  ELF \cite{Gray2008Viewpoint} & 12.00  & 59.50 & 74.50\\
  SalMatch \cite{Zhao2013SalMatch} & 30.16  & 62.50 & 75.60 \\
  MLF \cite{MidLevelFilter} & 29.11 & 65.20 & 79.90\\
  SCSP \cite{SimilaritySpatial} & $\mathbf{53.54}$  & $ \mathbf{91.49}$ &  $\mathbf{96.65}$\\
 Multi-channel \cite{Multi-channel-part} & 47.80  & 84.80 & 91.10\\
 NLML \cite{LocalMetric} & 42.30 & 85.23 & 94.25\\
 NullReid \cite{NullSpace-Reid} & 42.28  & 82.94 & 92.06\\
 DGDropout \cite{DGGropout} &38.40 & 86.60 & 90.89\\
 E-Metric \cite{E-Metric} & 40.90 & 85.05 & 92.00\\
 SI-CI \cite{SI-CI} & 35.80 & 83.50 & 97.07\\
 S-LSTM \cite{S-LSTM} &42.40 & 79.40 &-\\
 S-CNN \cite{GatedCNN} &37.80 & 77.40&-\\
 PIE \cite{PIE-reid} &18.10  & 38.92 & 49.40\\
\hline
   Ours &  49.11  & 87.66 &  93.47\\
  \hline
  \end{tabular}
  }
\end{table*}

\begin{table*}[t]
  \centering
  \caption{Rank-1, -10, -20 recognition rate of various methods on the CUHK03 data set (test person =100).}  \label{tab:cmc_cuhk03}
  {
  \begin{tabular}{l|c|c|c}
  \hline
\hline
    Method  & $R=1$  &   $R=10$  & $R=20$ \\
  \hline
   JointRe-id \cite{JointRe-id}  & $54.74$  & 91.50 & 97.31 \\
   FPNN \cite{FPNN} & $20.65$ & 51.32  & 83.06\\
   NullReid \cite{NullSpace-Reid}  & $58.90$  & $92.45$ & 96.30 \\
   SDALF \cite{Farenzena2010Person} & $5.60$  &36.09 & 51.96 \\
   eSDC \cite{eSDC} & $8.76$  &38.28 & 53.44 \\
   kLFDA \cite{Xiong2014Person} & 48.20  & 66.38 & 76.59\\
   LOMO+XQDA \cite{LOMOMetric} & 52.20  & 92.14 & 96.25\\
   PersonNet \cite{PersonNet} & 64.80  &94.92 & 98.20 \\
   DGDropout \cite{DGGropout} &72.60   & 93.50 & 96.70 \\
   E-Metric \cite{E-Metric} & 61.32 & 96.50 & 97.50 \\
   S-LSTM \cite{S-LSTM} & 57.30 & 88.30& 91.00\\
   S-CNN \cite{GatedCNN} & 61.80 & 89.30 & 92.20\\
   PIE \cite{PIE-reid} & 62.40 & 91.85 & 95.85\\
   SSM \cite{SSM} + GOG \cite{GOG} & 71.82 & 92.54 & 96.64\\
\hline
  Ours  &  \color{red}$\mathbf{73.23}$  & \color{red}$\mathbf{96.73}$ &   97.52\\
  \hline
  \end{tabular}
  }
\end{table*}

\begin{table*}[t]
  \centering
  \caption{Rank-1 and  mAP of various methods on the Market-1501 data set (test person =751).} \label{tab:cmc_market}
  {
  \begin{tabular}{l|c|c}
  \hline\hline
    Method  & $ R=1$  &  mAP \\
  \hline
   SDALF \cite{Farenzena2010Person} & 20.53 & 8.20\\
   eSDC \cite{eSDC} & 33.54 & 13.54\\
   kLFDA \cite{Xiong2014Person} & 44.37 &  23.14 \\
   XQDA \cite{LOMOMetric} & 43.79 & 22.22\\
   BoW \cite{Market1501} & 34.40 &  14.09\\
   SCSP \cite{SimilaritySpatial} & 51.90 & 26.35\\
   NullReid \cite{NullSpace-Reid}  & 61.02 & 35.68\\
   S-CNN \cite{GatedCNN} & 65.88 & 39.55\\
   SSM \cite{SSM}  & 82.21 & 68.80\\
   PIE \cite{PIE-reid} & 65.68 & 41.12 \\
   PIE \cite{PIE-reid} + KISSME \cite{Kostinger2012Large} & 79.33 &  55.95 \\
\hline
Ours & \color{red}$\mathbf{67.15}$ &40.24 \\
Ours + KISSME \cite{Kostinger2012Large} & 83.40 &  60.37\\
Ours + XQDA \cite{LOMOMetric} & \color{red}$\mathbf{86.15}$ &65.25 \\
  \hline
  \end{tabular}
  }
\end{table*}

\subsection{More Empirical Analysis on Our Approach}

\begin{figure}[t]
        \includegraphics[width=2.7in,height=2.3in]{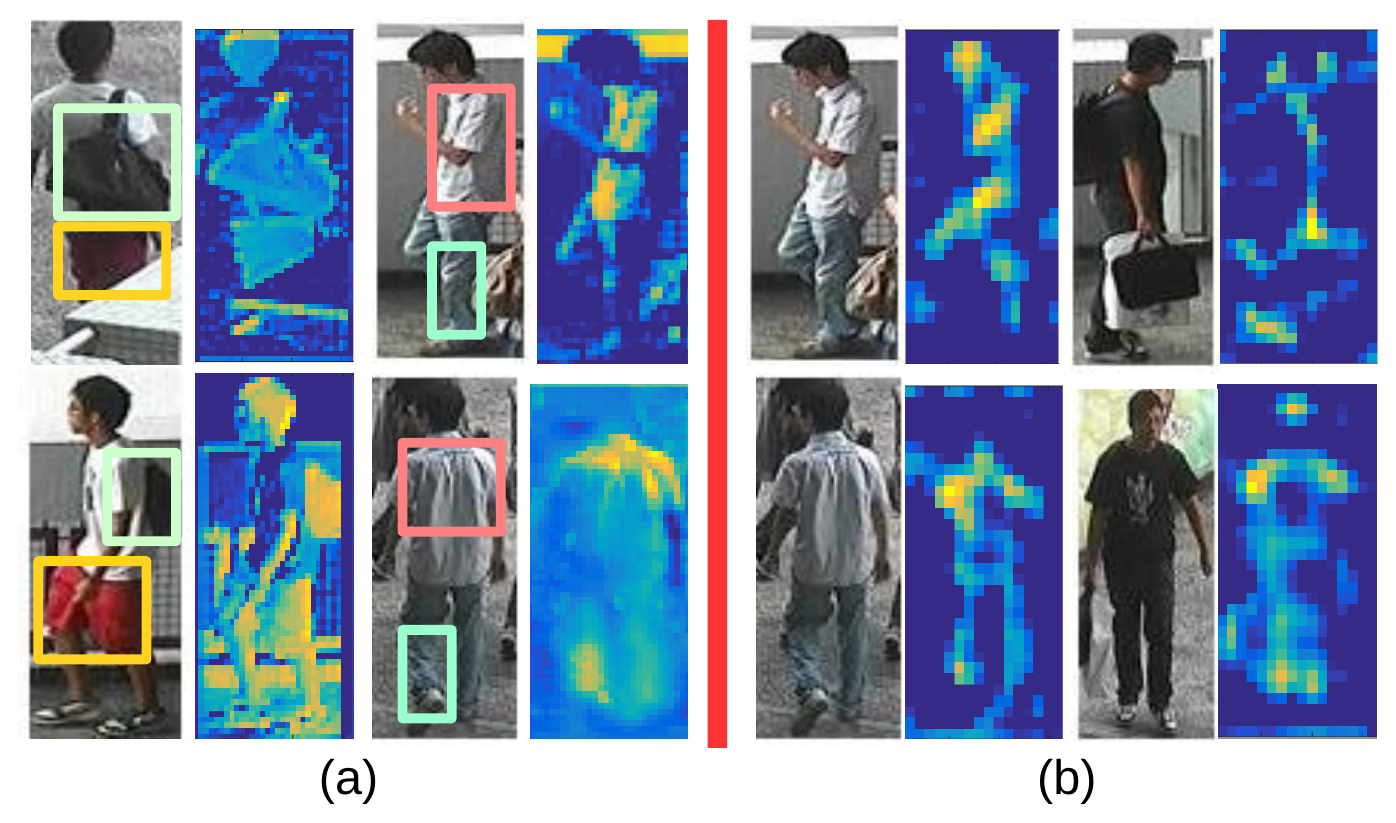}
        \includegraphics[width=2.5in,height=2.5in]{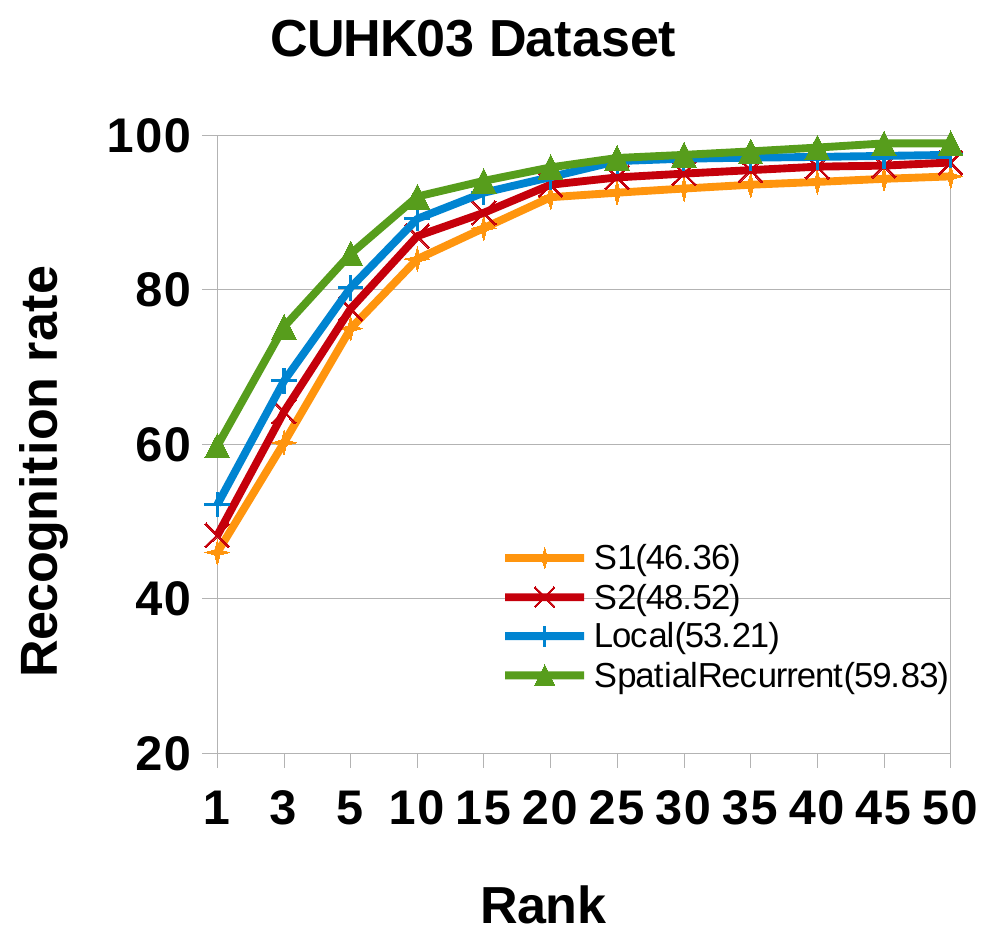}
    \caption{Left (a): Common distinct regions detected by integration gating computations. Left (b): High responses of spatially recurrent features. Raw RGB images in columns display the same identities observed by disjoint camera views. Right: CMC curves for comparison on individual/integrated local similarity and spatially recurrent measurement.}\label{fig:region_response}
\end{figure}

\paragraph{Understanding the Integrated Features}

The proposed model is motivated to seek common local subregions by joint deep embeddings. In this experiment, we analyze the network's specification into roles of localization and flexible appearance modeling in person re-id scenario when the network is initialized symmetrically and fine-tuned in a Siamese fashion. Fig.\ref{fig:region_response} (a) shows that the network tends to activate on highly semantic and common regions, such as back bag, and body parts. To evaluate the matching ability of these detected regions, we compute the local similarities corresponding to pairs of attentive regions. Suppose a pair of images has $R$ (we set $R=2$) detected regions, we extract their features, denoted as $x_a$ and $x_b$, respectively. Then, for each matching region $r$, we learn a similarity value: $S^r(x_a,x_b)=\langle \phi^r(x_a,x_b),W^r\rangle_F$, $r=\{1,\ldots,R\}$, where $\phi(x_a,x_b)=(x_a-x_b)(x_a-x_b)^T$, and $\langle\cdot,\cdot \rangle_F$ is the Frobenius inner product. Thus, $S^r(x_a-x_b)=(x_a-x_b)W^r(x_a-x_b)^T$ corresponds to the Mahalanobis distance. As shown in \cite{SimilaritySpatial}, local similarities are complementary and can be combined into an integrated measurement: $S^{Local}=\sum_{r=1}^R S^r(x_a,x_b)$. In Fig.\ref{fig:region_response}, $S1$, $S2$ and $Local$ show the matching rate of using detected regions independently, and an integrated similarity. We can see that (1) the integrated local similarity outperforms individual regions using deep features, and (2) upper body parts ($S2$) are more effective than lower body ($S1$). In future, we will study different matching properties regarding to body parts.

\paragraph{The Effect of Spatial Dependencies}

In this experiment, we investigate the effects of spatial manipulations in matching pedestrian images. Fig.\ref{fig:region_response} (b) shows the high responses from spatially recurrent features, and the evaluation of its matching rate is shown in Fig.\ref{fig:region_response}, where spatial recurrent achieves rank-1 rate $59.83\%$, a large margin over individual region matching and linear integrated local similarity.

\section{Conclusion}\label{sec:con}

In this paper, we present a novel deep spatially recurrent model for person re-identification that learns common local features with spatial manipulation to facilitate patch-level matching. The proposed scheme introduces an integration gating function to jointly embed pair-wised input images to discover the common patterns that are helpful in discriminating the positive pairs from negative ones for person re-id. To incorporate spatial dependencies into feature learning, stacked spatially recurrent pooling are embodied to make the learned representations spatially contextual. Comprehensive experiments show that our designed network achieves the superior performance on person re-identification. For the future work, we will continue to improve the models of part localization and matching with attention model.

\section*{Acknowledgement}

Junbin Gao's research was partially by Australian Research Council Discovery Projects funding scheme (Project No. DP140102270) and the University of Sydney Business School ARC Bridging Fund.

\bibliographystyle{named}
\bibliography{ijcai16}

\end{document}